\documentclass{article}
\usepackage{spconf,amsmath,graphicx}
	
\usepackage{booktabs,multirow,array}
\usepackage[table]{xcolor} 
\usepackage{caption}
\usepackage{tabularx}
\usepackage{caption}
\usepackage{rotating}
\usepackage{changepage}
\usepackage{subcaption}
\title{Simulating Patho-realistic Ultrasound Images using Deep Generative Networks with Adversarial Learning}
\name{Francis Tom, Debdoot Sheet }
\address{Indian Institute of Technology Kharagpur, India }
\begin{document}
%
\maketitle
\begin{abstract}
Ultrasound imaging makes use of backscattering of waves during their interaction with scatterers present in biological tissues. Simulation of synthetic ultrasound images is a challenging problem on account of inability to accurately model various factors of which some include intra-/inter scanline interference, transducer to surface coupling, artifacts on transducer elements, inhomogeneous shadowing and nonlinear attenuation. Current approaches typically solve wave space equations making them computationally expensive and slow to operate. We propose a generative adversarial network (GAN) inspired approach for fast simulation of patho-realistic ultrasound images. We apply the framework to intravascular ultrasound (IVUS) simulation. A stage 0 simulation performed using pseudo B-mode ultrasound image simulator yields speckle mapping of a digitally defined phantom. The stage I GAN subsequently refines them to preserve tissue specific speckle intensities. The stage II GAN further refines them to generate high resolution images with patho-realistic speckle profiles. We evaluate patho-realism of simulated images with a visual Turing test indicating an equivocal confusion in discriminating simulated from real. We also quantify the shift in tissue specific intensity distributions of the real and simulated images to prove their similarity.
\end{abstract}
\section{Introduction}
\label{sec:intro}

\begin{figure}[t]
  \centering
\includegraphics[width=0.47\textwidth]{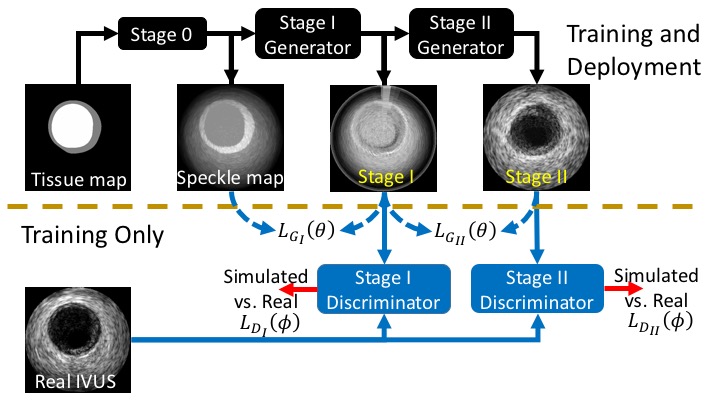}
\caption{Overview of the proposed framework for ultrasound simulation using stacked generative adversarial networks (GAN).}
\label{fig:res}
\end{figure}

Ultrasound images are formed by reflection of ultrasonic acoustic waves from body structures. This makes the task of simulating realistic appearing medical ultrasound images computationally complex~\cite{pipeline}. Such simulators could be used as a learning aid for doctors to be able to simulate and visualize co-morbid pathologies which occur rarely in reality. They can also simulate images to augmented datasets for supervised learning algorithms. Fig.~\ref{fig:res} summarizes the contribution of this paper where we present a generative adversarial network (GAN) based framework for fast and realistic simulation of ultrasound images. The GAN framework~\cite{GAN} consists of two neural networks: a \emph{generator} and a \emph{discriminator}. In an adversarial learning strategy, the \emph{generator} learns while attempting to generate realistic ultrasound images. The \emph{discriminator} simultaneously learns while attempting to discriminate between true images and those simulated by the \emph{generator}. This learning rule manifests as a two player mini-max game. Our formulation is elucidated in Sec.~\ref{sec:method}.

The rest of the paper has Sec.~\ref{sec:priorart} describing the prior art of ultrasound image simulation. Sec.~\ref{sec:expt} presents the results of our experiments on the publicly available border detection in IVUS challenge\footnote{http://www.cvc.uab.es/IVUSchallenge2011/dataset.html} dataset along with discussion. Sec.~\ref{sec:conc} presents the conclusions of the work.

\section{Prior Art}
\label{sec:priorart}

The different approaches for simulating ultrasound images can broadly be grouped as employing (i) numerical modeling of ultrasonic wave propagation or (ii) image convolution operation based. An early approach employed convolutional operators leveraging the linearity, separability and space invariance properties of the point spread function (PSF) of imaging system to simulate pseudo B-mode images~\cite{Bamber}. Numerical modeling based approaches have employed ray-based model to simulate ultrasound from volumetric CT scans~\cite{GPU}, and with the Born approximation~\cite{BORNPAPER} have also simulated images using histopathology based digital phantoms employing wave equations~\cite{RAY, DS, MICCAI}. Further Rayleigh scattering model~\cite{RAYLEIGH}, physical reflection, transmission and absorption have also been included in simulation~\cite{REAL_IVUS}. Commercially available solutions like Field II simulator\footnote{http://field-ii.dk/} have used a finite element model for the purpose. These simulators that model the tissue as a collection of point scatterers fail to model appearance of artifacts, and the high computational complexity challenge their deployment. Recent approaches employ spatially conditioned GAN~\cite{SGAN}, yet are limited in resolution and the appearance of unrealistic artefacts in the generated images.

\section{Methodology}
\label{sec:method}

Our framework employs (1) simulation of ultrasound images from the tissue echogenicity maps using a physics based simulator, (2) generation of low resolution images from the earlier simulated images by learning Stage I GAN, (3) generation of high resolution patho-realistic images from the low resolution earlier stage by a Stage II GAN. Stage I GAN does the structural and intensity mapping of the speckle maps generated by the physics based simulator. Stage II GAN models patho-realistic speckle PSF and also corrects any defects in the images simulated by the stage I GAN. Generation of high resolution images by GANs is difficult due to training instability~\cite{GAN_theory}. Hence the complex task of ultrasound simulation is decomposed into sub-problems. 

\subsection{Stage 0 simulation from tissue echogenicity map}
\label{sec:method:subsec:gan0}

The ground truth label of lumen and external elastic lamina boundary contour available in the dataset is used to generate the tissue echogenicity map (Fig.~\ref{fig:PolarTissue map}). The initial simulation is performed by a pseudo B-mode ultrasound image simulator\footnote{https://in.mathworks.com/matlabcentral/fileexchange/34199}~\cite{Bamber, SRAD} that assumes a linear and space invariant point spread function (PSF). Image is formed with assumption of wave propagating vertically along the echogenicity map. The images are generated in the polar domain (Fig.~\ref{fig:PolarPseudo B-Mode}).

\subsection{Stage I GAN}
\label{sec:method:subsec:gan1}

\begin{figure*}[t]
 \centering
    \begin{subfigure}[b]{0.18\textwidth}
        \centering
        \includegraphics[width=\textwidth]{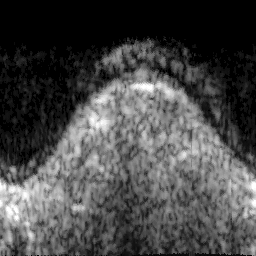}
        \caption{Real image}
        \label{fig:PolarReal image}
    \end{subfigure}
        \hfill
    \begin{subfigure}[b]{0.18\textwidth}
        \centering
        \includegraphics[width=\textwidth]{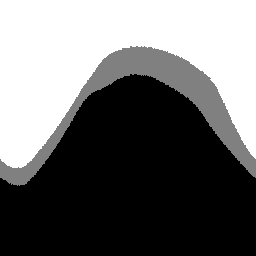}
        \caption{Tissue map}
        \label{fig:PolarTissue map}
    \end{subfigure}
    \hfill
    \begin{subfigure}[b]{0.18\textwidth}
        \centering
        \includegraphics[width=\textwidth]{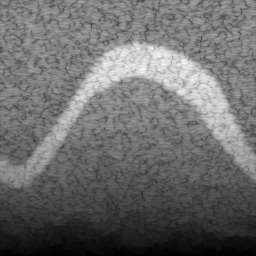}
        \caption{Pseudo B-Mode}
        \label{fig:PolarPseudo B-Mode}
    \end{subfigure}
    \hfill
    \begin{subfigure}[b]{0.18\textwidth}
        \centering
        \includegraphics[width=\textwidth]{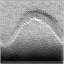}
        \caption{Stage I GAN}
        \label{fig:PolarStage I GAN}
    \end{subfigure}
    \hfill
        \begin{subfigure}[b]{0.18\textwidth}
        \centering
        \includegraphics[width=\textwidth]{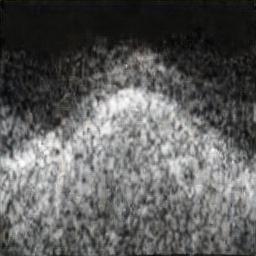}
        \caption{Stage II GAN}
        \label{fig:PolarStage II GAN}
    \end{subfigure}

    \centering
    \begin{subfigure}[b]{0.18\textwidth}
        \centering
        \includegraphics[width=\textwidth]{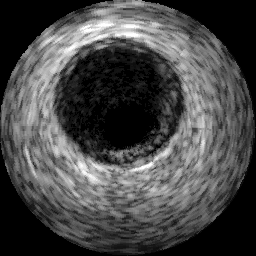}
        \caption{Real image}
        \label{fig:Real image}
    \end{subfigure}
    \hfill
    \begin{subfigure}[b]{0.18\textwidth}
        \centering
        \includegraphics[width=\textwidth]{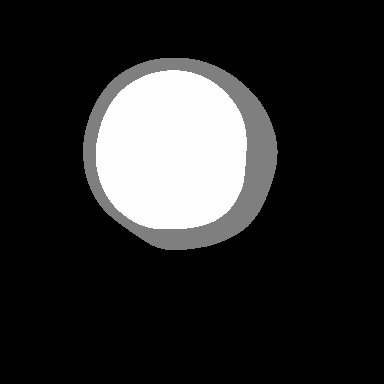}
        \caption{Tissue map}
        \label{fig:Tissue map}
    \end{subfigure}
    \hfill
    \begin{subfigure}[b]{0.18\textwidth}
        \centering
        \includegraphics[width=\textwidth]{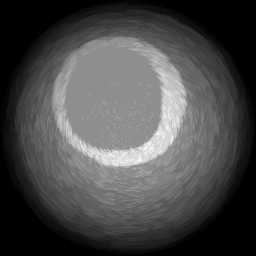}
        \caption{Pseudo B-Mode}
        \label{fig:Pseudo B-Mode}
    \end{subfigure}
    \hfill
        \begin{subfigure}[b]{0.18\textwidth}
        \centering
        \includegraphics[width=\textwidth]{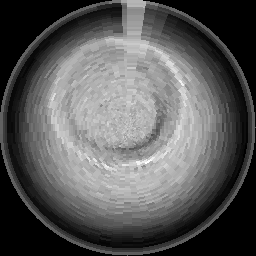}
        \caption{Stage I GAN}
        \label{fig:Stage I GAN}
    \end{subfigure}
    \hfill
    \begin{subfigure}[b]{0.18\textwidth}
        \centering
        \includegraphics[width=\textwidth]{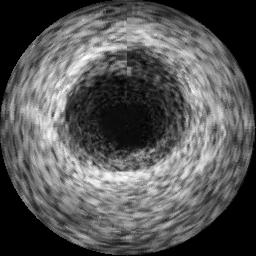}
        \caption{Stage II GAN}
        \label{fig:Stage II GAN}
    \end{subfigure}
         \caption{Comparison of images obtained at different stages of the proposed framework. The first row displays images in the Polar domain and the second row displays the corresponding figures in the Cartesian domain.}
    \label{fig2}
\end{figure*}

Refinement of the output of the pseudo B-mode simulator is performed by a two stage stacked GAN~\cite{stackgan}. The Stage I GAN is trained to generate a low resolution image of $64\times64$ px. size in the polar domain using the simulated + unsupervised learning approach~\cite{simgan}.  This involves learning of a refiner network $G_I({\mathbf x})$ to map the speckle map generated by the pseudo B-mode simulator ${\mathbf x}$, with ${\mathbf \theta}$ as the function parameters. The discriminator network $(D_{I_{\mathbf \phi}})$ is trained to classify whether the images are real or refined, where ${\mathbf \phi}$ denotes the discriminator network's parameters. The refined image is denoted as $\overset{\sim}{{\mathbf x}}$ (Fig.~\ref{fig:PolarStage I GAN}). The set of real images ${\mathbf{y}_i} \in \Upsilon$ (Fig.~\ref{fig:PolarReal image}) are also used for learning the parameters. The loss function consists of a realism loss term ($l_{real}$) and an adversarial loss term. The realism loss term ensures that the ground truth label remains intact in the low resolution refined image which is used to condition the Stage II GAN. During training of the Stage I GAN, the discriminator loss $(L_{D_{I}})$ and generator loss $(L_{G_{I}})$ are minimized alternately where,

 \begin{equation}
 L_{G_I}({\mathbf \theta}) = -\sum_{i}^{}\log (1-{D_I}_{{\mathbf \phi}}({G_I}_{{\mathbf \theta}}({\mathbf{x}_i}))+\lambda l_{reg}({\mathbf \theta};{\mathbf{x}_i})
 \end{equation}
 
 \begin{multline}
L_{D_I}({\mathbf \phi}) = -\sum_{i}^{} \log ({D_I}_{{\mathbf \phi}}({G_I}_{\theta}({\mathbf{x}_i}))) \\ -\sum_{j}^{}\log (1-{D_I}_{{\mathbf \phi}}({\mathbf{y}_j}))
\end{multline}

\noindent Self-regularization here minimizes per-pixel difference between the refined image and the synthetic image as 

\begin{equation} l_{reg} = 
\lVert \overset{\sim}{{\mathbf x}}-{\mathbf x} \rVert_1\end{equation}

\noindent A buffer of refined images generated over the previous steps was used to improve the realism of the artifacts in the refined images as in~\cite{simgan}.

\textbf{Model architecture:}
 The Stage I GAN is a residual network~\cite{resnet} with $4$ residual blocks. The simulated image from the pseudo B-mode simulator of size $64\times64$ is first passed through the \emph{generator} with convolution layer with $3\times3$ sized kernel and $64$ feature maps on output. The refined image is generated by a convolution layer with $1\times1$ sized kernel. The \emph{discriminator} comprises of $5$ convolution layers and $2$ max pooling layers as in~\cite{simgan}.

\subsection{Stage II GAN}
\label{sec:method:subsec:gan2}

The generator of the Stage II GAN accepts the low resolution images from the Stage I GAN generator and generates high resolution images adding realistic artifacts. While training the Stage II GAN, the discriminator loss ($L_{D_{II}}$) and generator loss ($L_{G_{II}}$) are minimized alternately where,

 \begin{equation}
 L_{G_{II}}({\mathbf \theta}) = -\sum_{i}^{}\log (1-{D_{II}}_{{\mathbf \phi}}({G_{II}}_{{\mathbf \theta}}({G_I}({\mathbf{x}_i})))
 \end{equation}
 
 \begin{multline}
L_{D_{II}}({\mathbf \phi}) = -\sum_{i}^{} \log ({D_{II}}_{{\mathbf \phi}}({G_{II}}_{{\mathbf\theta}}({G_I}({\mathbf{x}_i}))) \\ -\sum_{j}^{}\log (1-{D_{II}}_{{\mathbf \phi}}({\mathbf{y}_j}))
\end{multline}

\noindent The high resolution images generated are in polar domain (Fig.~\ref{fig:PolarStage II GAN}) and scan converted to cartesian coordinate domain (Fig.~\ref{fig:Stage II GAN}) for visual inspection.

\textbf{Model architecture:}
The Stage II GAN \emph{generator} consists of a downsampling block that comprises of two maxpooling layers to downsample the $64\times 64$ image by a factor of four. The feature maps are then fed to residual blocks, followed by upsampling the image to $256\times 256$. The \emph{discriminator} consists of downsampling blocks that reduce the dimension to $4\times 4$ followed by a $1\times1$ convolution layer and a fully connected layer to generate the prediction.

\section{Experiments, Results and Discussion}
\label{sec:expt}

In this section, experimental validation of our proposed framework is presented. Images from the dataset released as part of the border detection in IVUS challenge\footnote{http://www.cvc.uab.es/IVUSchallenge2011/dataset.html} were used in our work. The images were obtained from clinical recordings using a 20 MHz probe. Of the $2,175$ available images, $435$ had been manually annotated. We used $2,025$ real images for training (Patient Id. 1-9) and $150$ images were held out for testing (Patient Id. 10 ). The ground truth labeled dataset containing $435$ images were augmented by rotating each image by $30^{\circ}$  over $12$ steps. Each such image was then warped by translating up and down by $2\%$ in the polar domain to yield an augmented dataset of $15,660$ tissue maps to be processed with the pseudo B-mode simulator. 

The Stage I GAN was trained with a learning rate of $0.001$ with the Adam optimizer over $20$ epochs with a batch size of $512$. While training the Stage II GAN, the weights of Stage I GAN were held fixed. Training was done with the Adam optimizer with an initial learning rate of $0.0002$, and a learning rate decay of $0.5$ per $100$ epochs over $1,200$ epochs with a batch size of $64$.

Images obtained at different stages of the pipeline are provided in Fig.~\ref{fig2}. The Stage I GAN generates low resolution (64 x 64) images  and are blurry with many defects. As shown, the Stage II GAN refines the image generated by Stage I GAN, adding patho-realistic details at $4\times$ higher resolution.

\textbf{Qualitative evaluation of the refined images:}
In order to evaluate the visual quality of the adversarially refined images, we conduct a Visual Turing Test (VTT). A set of $20$ real images were randomly selected from the $150$ images in test set. Each real image was paired with simulated image and presented to $13$ evaluators who had experience in clinical or research ultrasonography. Each evaluator was presented with an independently drawn $20$ pairs and they had to identify the real image in each pair. Cumulatively they were able to correctly identify $147$ real IVUS out of $260$ present with $56\%$ chance of correct identification.

\textbf{Quantitative evaluation of the refined images:}
The histograms of the real and adversarially refined images in the three regions i.e., lumen, media and externa for $30$ randomly chosen annotated images were obtained and the probability mass functions were computed from the histograms. A random walk based segmentation algorithm~\cite{random} was used for the annotation of the refined images. The Jensen-Shannon (JS) divergence between the probability mass functions of the three regions in real vs. simulated images were computed and summarized in Table.~\ref{quant1}. It is observed that result of Stage II GAN are realistically closer to real IVUS as compared to Stage 0 results. The pairwise JS-divergence between lumen and media, media and externa and lumen and externa were computed and summarized in Table.~\ref{quant2}. The divergence between pair of tissue classes in Stage II GAN are also realistically closer to real IVUS.

 \begin{table}[t]
\centering
\caption{JS-divergence between speckle distribution of different regions in simulated vs. real IVUS.}
\label{quant1}
\begin{tabular}{| l | c | c | c |}
\hline
      & lumen & media & externa \\ \hline
Stage II GAN & 0.0458       & 0.1159       & 0.3343     \\ \hline
Stage 0 & 0.0957       & 0.5245      & 0.6685     \\ \hline

\end{tabular}
\end{table}

\begin{table}[t]
\centering
\caption{JS-divergence between pair of tissue types in real and simulated IVUS.}
\label{quant2}
\begin{tabular}{| l | c | c | c |}
\hline
 & lumen-media & media-ext. & lumen-ext. \\ \hline
Real & 0.2843       & 0.1460       & 0.2601     \\ \hline
Stage II GAN & 0.1471       & 0.1413       & 0.3394     \\ \hline
\end{tabular}
\end{table}

\section{Conclusion}
\label{sec:conc}

Here we have proposed a stacked GAN based framework for the fast simulation of patho-realistic ultrasound images using a two stage GAN architecture to refine images synthesized from an initial simulation performed with a pseudo B-model ultrasound image generator. Once training is completed, images are generated by providing a tissue map. The quality of the simulated images was evaluated through a visual Turing test that evoked equivocal visual response across experienced raters with ~$50\%$ chance of identifying the real from simulated IVUS. The quality of the simulated images was quantitatively evaluated using JS divergence between the distributions of the real and simulated images. Similarity of the real and simulated images is also quantified by computing the shift in tissue specific speckle intensity distributions. Conclusively these evaluations substantiate ability of our approach to simulate patho-realistic IVUS images, converging closer to real appearance as compared to prior-art, while simulating an image in under $10$ ms during deployment.

\small
\bibliographystyle{IEEEbib}
\bibliography{egbib}

\end{document}